\documentclass[twoside,11pt]{article}

\usepackage{jmlr2e}
\usepackage{adjustbox}
\usepackage{epsfig} % for postscript graphics files 
\usepackage{graphicx}
\usepackage{amsmath} % assumes amsmath package installed
\usepackage{amssymb}  % assumes amsmath package installed
\usepackage{algorithm}
\usepackage{algpseudocode}
\usepackage{ctable}
\usepackage{multirow}
\usepackage{epstopdf}
\usepackage{color, soul}
\usepackage{colortbl}
\usepackage{tabularx}
\usepackage{pifont}
\newcommand{\vect}[1]{\boldsymbol{#1}}

\definecolor{tabrow}{rgb}{0.88,0.88,0.88}
\definecolor{Gray}{gray}{0.8}

\usepackage[utf8]{inputenc} % allow utf-8 input
\usepackage[T1]{fontenc}    % use 8-bit T1 fonts

\ShortHeadings{Meta-Weighted Gaussian Process Experts}{Rudovic et al.}
\begin{document}

\title{Meta-Weighted Gaussian Process Experts \\for Personalized Forecasting of AD Cognitive Changes}
\author{\name Ognjen (Oggi) Rudovic$^1$ \email{orudovic@mit.edu} 
       \AND
       \name Yuria Utsumi$^1$ \email{yutsumi@mit.edu} \AND
       \name Ricardo Guerrero$^2$ \email{reg09@imperial.ac.uk} 
       \AND
       \name Kelly Peterson$^1$ \email{kellypet@mit.edu} 
       \AND
       \name Daniel Rueckert$^2$ \email{d.rueckert@imperial.ac.uk}  
       \AND
       \name Rosalind W. Picard$^1$ \email{roz@mit.edu} \\\\
       \addr{$^1$Massachusetts Institute of Technology, $^2$Imperial College London\\}}

\maketitle

\begin{abstract}
We introduce a novel personalized Gaussian Process Experts (pGPE) model for predicting per-subject ADAS-Cog13 cognitive scores -- a significant predictor of Alzheimer's Disease (AD) in the cognitive domain -- over the future 6, 12, 18, and 24 months. We start by training a population-level model using multi-modal data from previously seen subjects using a base Gaussian Process (GP) regression. Then, we personalize this model by adapting the base GP sequentially over time to a new (target) subject using domain adaptive GPs, and also by training subject-specific GP. While we show that these models achieve improved performance when selectively applied to the forecasting task (one performs better than the other on different subjects/visits), the average performance per model is suboptimal. To this end, we used the notion of meta learning in the proposed pGPE to design a regression-based weighting of these expert models, where the expert weights are optimized for each subject and his/her future visit. The results on a cohort of subjects from the ADNI dataset show that this newly introduced personalized weighting of the expert models leads to large improvements in accurately forecasting future ADAS-Cog13 scores and their fine-grained changes associated with the AD progression. This approach has potential to help identify at-risk patients early and improve the construction of clinical trials for AD.

\end{abstract}

\section{Introduction}

Alzheimer's disease (AD) is the most common form of dementia, usually associated with the elderly population (over 65 years of age). AD had a worldwide prevalence of around 33.9 million cases by 2011, with predictions suggesting an increase to about 100 million by 2050
\citep{barnes2011}. Recently, the view on AD diagnosis has shifted towards a more dynamic process in which clinical and pathological markers evolve gradually before diagnostic criteria are met. The AD Assessment Scale-cognition sub-scale (ADAS-Cog)~\citep{mohs1997development} is the most widely used general cognitive measure in clinical trials of AD~\citep{skinner2012alzheimer}. While it was developed as an outcome measure for dementia interventions, its primary purpose was to be an index of global cognition in response to antidementia therapies. The ADAS-Cog assesses multiple cognitive domains including memory, language, praxis, and orientation~\citep{skinner2012alzheimer}. Because ADAS-Cog has proven important for target clinical assessments, in this paper we focus on a machine learning method that can successfully forecast the future values of this score for target subjects. Specifically, we use the modified ADAS-Cog 13-item scale~\citep{mohs1997development}, which is scored on a 85 point scale, where higher scores indicate greater severity.

\begin{figure}[t]
\centering
\caption{{\small {\bf Personalized GP Experts (pGPE):} The population model is first trained using all visits data of $N$ training subjects ($x^{TR}$ are the input features and $y^{TR}$ are the corresponding ADAS-Cog13 scores), where the time difference between two visits is 6 months. The model personalization to the target (previously unseen) subject ($N+1$) is then achieved by sequentially adapting the model predictions of the future ADAS-Cog13 scores $y_{t+1:t+4}$ (using the posterior distribution of GPs - $f_{GP}$), informed by the visits data up to time step $t$, in the personalized GP (pGP) model. We also train the subject-specific model - target GP (tGP) - using only the past data of the target subject. The shaded fields in the output vector represent the time points for which we aim to predict ADAS-Cog13 scores. The proposed pGPE performs a regression-based weighting of the experts' pGP and tGP using meta-weights ($\alpha$), estimated using a GP regression model (GP). In this way, the optimal weights are trained for each subject and his/her future visit, leading to large improvements over the individual experts (pGP and tGP) in the target task.}}
\label{overview}
\includegraphics[scale=.3]{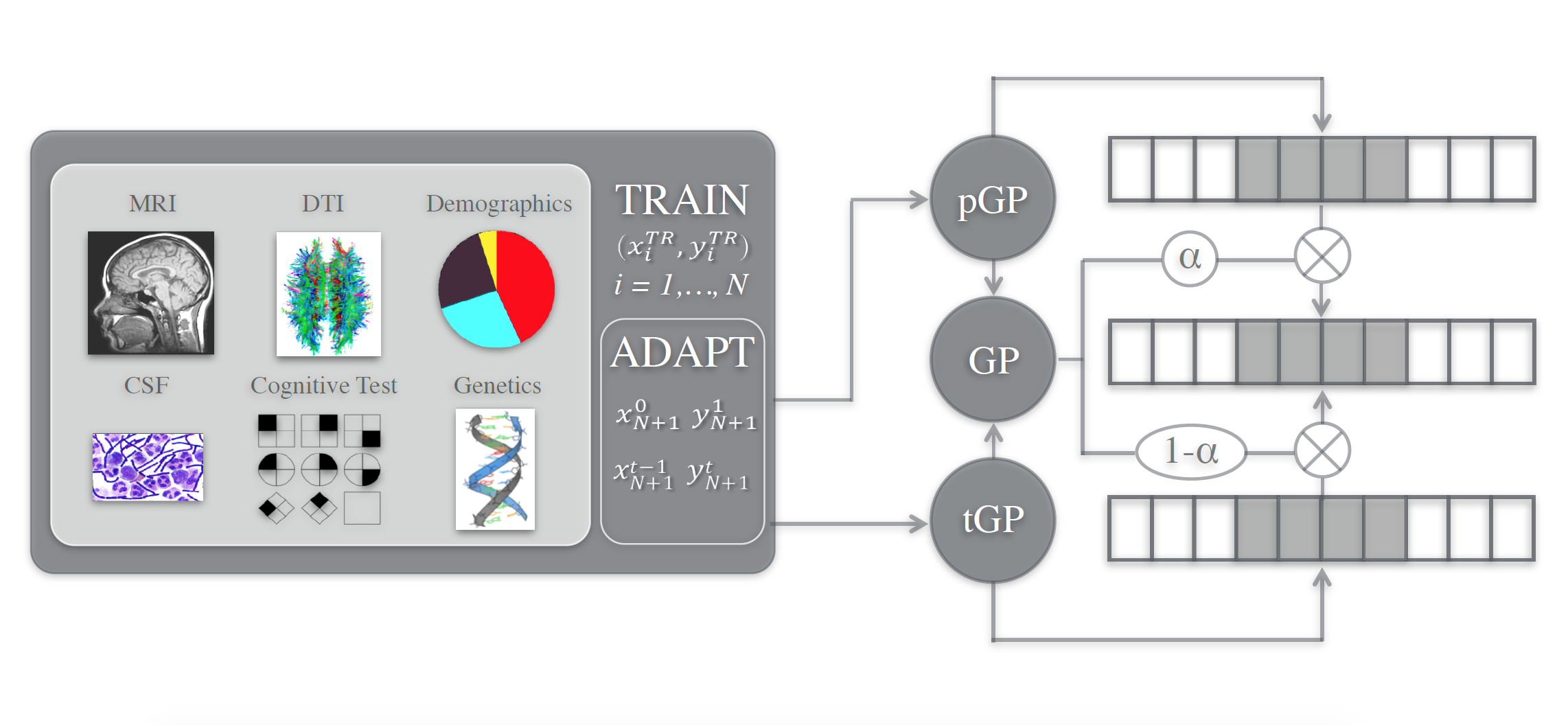}
\end{figure}
One of the main challenges in the clinical assessment of subjects at risk of developing AD is the ability to accurately predict subjects' future cognitive scores. Such predictions can play an important role in subject selection during the design of AD clinical trials. For this, an automated approach that could forecast future cognitive scores, such as ADAS-Cog13, would be of great value during assessment procedures, and could potentially improve clinical trial design and early detection of at-risk subjects. For example, out of hundreds of clinical trials, costing billions of dollars, fewer than $1\%$ have proceeded to the regulatory approval stage and none have managed to prove a disease-modifying effect \citep{tadpole2017,cummings2006}. 
Previously, it has been suggested that the low success rate of pharmaceutical AD clinical trials could be due to the inclusion of study populations that were too heterogeneous \citep{falahati2014}.
More successes depend on an improved ability to accurately identify subjects at early ages of the disease where treatments are most likely to be effective. Thus, developing models with improved ability to automatically predict subjects' future AD-related metrics indicating disease progression -- and to do so as early as possible, especially before the emergence of clinical symptoms -- is an important step towards this end. Furthermore, accurate prediction of symptom onset in the time window of 6 to 24 months is critical to participant selection and formation of clinical trials. Thus, having access to accurate future estimates of the progression of cognitive scores such as ADAS-Cog13 within this time frame is of great importance. 

In this paper, data from the Alzheimer’s Disease Neuroimaging Initiative (ADNI)~\citep{weiner2017} were used, and, specifically, the dataset processed for the TADPOLE Challenge \citep{tadpole2017}. These data are highly heterogeneous and multi-modal, and include imaging (MRI, PET), cognitive scores, CSF biomarkers, genetics, and demographics (e.g. age, gender, race). Although the heterogeneous nature of this dataset lends itself to building powerful, informative multi-modal models, the dataset itself is very sparse, with different combinations of features missing for different subjects. Partial records are also missing for roughly $80\%$ of the subjects \citep{campos2015}. Moreover, given the wide variability in available data per subject, inherent per-person differences, and the slowly changing nature of the disease, accurate forecasting of cognitive decline and related measures of disease progression is a significant and difficult challenge. To tackle these challenges, we focus on machine learning models that can easily adapt to each subject when forecasting his/her cognitive scores. More specifically, we investigate the effects of the model personalization for the forecasting task using the framework of Gaussian Processes (GP)~\citep{rasmussen2006gaussian}. This non-parametric probabilistic framework offers great modeling flexibility for not only predicting future cognitive scores but also uncertainty in their predictions. This, in turn, provides a principled approach for model adaptation to target subjects via posterior distribution of GP~\citep{peterson2017personalized}. Moreover, GP are well-suited for forecasting tasks where data are noisy time series with many missing values~\citep{futoma2018gaussian}, as in ADNI.

%Oggi: change the order -- contributions clearer

To this end, we used a cohort of subjects from the ADNI dataset (see Sec.~\ref{exp}) to predict future ADAS-Cog13 scores of target subjects using data from each subject’s previous visits. A subject's visit is defined as data collected at a single time point (a subject visit) during ADNI. We start by training a population-level model using multi-modal data of previously-seen (source) subjects using the base GP regression (sGP). Then, this model is adapted sequentially over time to a new (target) subject using the notion of domain adaptive GPs~\citep{liu2015bayesian,eleftheriadis2017}. We extend this personalization approach, denoted as pGP~\citep{peterson2017personalized}, for the forecasting task of predicting the ADAS-Cog13 scores at 6, 12, 18 and 24 months in the future. We also compare this approach with the subject-specific GP model, trained only on the data of the target subject (tGP). While we show that these models achieve good performance when selectively applied to the forecasting task (one performs better than the other for certain subjects/visits), their individual (average) performance is suboptimal. 
To this end, we propose a novel weighting scheme for the GP experts based on the notion of meta-learning~\citep{vanschoren2018meta}. The main idea of meta-learning is to observe how different machine learning approaches perform on a wide range of learning tasks, and then learn from this experience, or meta-data, to learn new tasks more effectively. Following this approach, we designed a regression-based weighting of these expert models in the proposed personalized GP experts (pGPE) model, where the expert weights are optimized for each subject and his/her future visit using data-derived meta-features (see Sec.~\ref{pgpe}). Our results show that this newly introduced personalized weighting of the expert models leads to large improvements in accurately forecasting future ADAS-Cog13 scores for each target subject, while also outperforming traditional weighting schemes for expert models (e.g. by simply averaging the model predictions or using the GP-variance-based weighting), and the standard models for time-series forecasting tasks.   

\section{Related Work}
We briefly review here the related work on forecasting of cognitive scores and clinical status for AD assessment, with the focus on the ADNI dataset. Most existing approaches, e.g.,~\citep{schmidt-richberg16,guerrero2016,gavidia2017}, focus on modeling subjects based on their clinical status (CS): cognitively normal (CN), mild cognitive impairment (MCI) and Alzheimer's Disease (AD). Additionally, the majority of these model biomarkers at the population level; for instance, estimating typical trajectories of markers over the full course of the disease to estimate current disease progress and progression rate \citep{schmidt-richberg16,guerrero2016}. Guerrero et al.~\citep{guerrero2016} used mixed effects modeling to derive global and individual biomarker trajectories for a training population, which was later used to instantiate subject-specific models for unseen subjects. Some of the modeling techniques \citep{guerrero2016,schmidt_richberg15_a,schmidt-richberg16} require cohorts with known disease onset and are prone to bias due to the uncertainty of the conversion time.

When it comes to forecasting clinical status, several authors attempted predicting target scores for longer time windows. For example, several recent works have explored the use of multiple, multi-modal predictors in combination with various machine learning techniques (e.g. SVMs, neural nets, GPs, etc.) to predict conversion from MCI to AD for various future time periods, e.g. from 1-5 years after baseline assessment~\citep{long2017,minhas2017}. Likewise, the BrainAGE framework has been proposed for predicting MCI-AD conversion within 3 years of follow-up \citep{gaser2013}. In addition, to predict conversion within time windows of up to 2 years (short-term converter) and 2-4 years (long-term converter), \citep{pereira2017} proposed a stepwise learning approach, where the learned model first predicts whether a subject converts to dementia, or remains stable, and then predicts the more likely progression window (short-term or long-term conversion). More recently, \citep{grassi2018a} evaluated different machine learning models in order to develop an algorithm for a 3-year prediction of conversion to AD in MCI and PreMCI subjects based only on non-invasive and effectively collectible predictors.

Most of these works attempted forecasting the changes in subjects' CS, which deals with a limited number of future outcomes (i.e. either binary or a three-class). By contrast, we aim to forecast ADAS-Cog13, defined on a more fine-grained scale (85 levels), which poses a more challenging machine learning problem. While recent work investigated forecasting of ADAS-Cog13~\citep{peterson2017personalized}, along with other cognitive scores from the TADPOLE challenge, using the GP framework, their forecasting horizon was limited as it only predicted one time step ahead (6 months). In this work, we attempt forecasting of ADAS-Cog13 up to 24 months ahead. While recent work in~\citep{utsumi2018pgp} has shown that a simple combination of personalized GP models leads to better prediction accuracy of the target cognitive scores, to the best of our knowledge, no prior work has proposed a principled approach for combining this expert knowledge. In this paper, we do so using the notion of meta learning to design an optimal weighting scheme for personalized GP models.

\section{Methodology}
\label{methods}
{\bf Notation.} We consider a supervised setting, where $\vect{X} =\{\vect{X}^{\bf(s)},\vect{X}^{\bf(t)}\}$ 
represent the input features to our model, and superscripts $\bf l=\{s,t\}$ refer to the data of source (i.e., training) $\bf (s)$  and target $\bf (t)$ subjects, respectively. Similarly, the ADAS-Cog13 scores we aim to forecast are stored in $\vect{Y} =\{\vect{Y}^{\bf (s)},\vect{Y}^{\bf(t)}\}$. More specifically, $\vect{X}^{\bf(l)} = \{\vect{x}^{\bf(l)}_{n}\}_{n=1}^{N_{\bf l}}$, where $N_{\bf s}$ and $N_{\bf t}$ are the numbers of the source and target subjects, respectively. Furthermore, $\vect{x}=[\vect{x}_1,\dots,\vect{x}_t,\dots,\vect{x}_{\vect{T}}]$ are the input features (predictors) for each time step of the subject's data\footnote{The subscript $t$ refers to time and is different from superscript $\bf(t)$ that refers to the target subject.}, and $\vect{T}\leq13$ is the maximum number of visits per subject over the period of 10 years, which varies largely per subject. Each $\vect{x}_t$ is a vector containing the features as $\vect{x}_t=[x_t^{m_1},\dots,x_t^{m_6},y_t]^T$, where $T$ is the transpose operation, and $m_{i=1,\dots,6}$ denotes the features from each data modality. We also add the current cognitive score $y_t\in\text{(0-85)}$ as a predictor since we found that it is a strong predictor of the future scores. The scores that we forecast are stored in $\vect{Y}^{\bf(l)} = \{\vect{y}^{\bf(l)}_{n}\}_{n=1}^{N_{\bf l}}$, and as $\vect{y}=[\vect{y}_1,\dots,\vect{y}_t,\dots,\vect{y}_{\vect{T}}]$. Here, each $\vect{y}_t$ contains the scores in the forecasting window of four steps ahead, and is given by $\vect{y}_t=[y_{t+1},\dots,y_{t+4}]^T$. Thus, the goal of the forecasting models proposed here is to learn an efficient mapping: $\vect{X}\rightarrow\vect{Y}$. As many subjects in ADNI missed certain visits and not all biomarkers were recorded at every visit, we fill in subjects' missing values using their nearest available past visit; however, no data of future visits are used.\footnote{To tackle this, more advanced approach based on auto-encoders can be used, e.g., see~\citep{campos2015}, and we leave this for the future work.}

\subsection{Population-level GPs} 
We first build the population-level forecasting model using data from the source subjects $\{\vect{X}^{\bf(s)},\vect{Y}^{\bf(s)}\}$.
To this end, we use the GP framework~\citep{rasmussen2006gaussian} to train the following forecasting function:
\begin{equation}
 \vect{y}_{t}^{\bf (s)} = f(\vect{x}_{t}^{\bf (s)}) + \epsilon_{t}^{\bf (s)},
\end{equation}
where $\epsilon_t^{\bf(s)}\sim \mathcal{N}(0, \sigma^2_{\bf s})$ is i.i.d. additive Gaussian noise, which variance $\sigma^2_{\bf s}$ is estimated from the training data. In this framework, central to learning of the forecasting function $f(\cdot)$ is the GP prior that is placed over the function space, leading to the marginal likelihood of our data $p(\vect{Y}^{\bf(s)}|\vect{X}^{\bf(s)}) = \mathcal{N}(\vect{Y}^{\bf(s)}|\vect{0}, \vect{K}^{\bf(s)}+\sigma_{\bf s}^2\vect{I})$, where the elements of the kernel matrix are given by $\vect{K}^{\bf(s)} = \vect{k}(\vect{X}^{\bf(s)},\vect{X}^{\bf(s)})$ and $\vect{I}$ is the identity matrix. The choice of the kernel function $k(\cdot,\cdot)$ is at the heart of GPs as it encodes the relationships in our data. To this end, we use the standard radial basis function (RBF) kernel. Specifically, we investigate two types of the RBF kernel: isotropic (iso) and automatic relevance determination (ard)~\citep{rasmussen2006gaussian}. The latter assigns different weights (length-scales) to each feature (in our case, each data modality), effectively doing feature selection. 

The kernel parameters $\vect{\theta}$ are optimized by minimizing the negative log-marginal likelihood: $-\log p(\vect{Y}^{\bf(s)}|\vect{X}^{\bf(s)}, \vect{\theta})$ using conjugate gradient method. Then, given the data from the visit at time $t$ of a {\it new} subject, $\vect{x}_\ast=\{\vect{x}_{t},\vect{y}_{t}\}$, the GP predictive distribution provides the mean and variance forecasts of the cognitive scores $\vect{y}_\ast$ as:
\begin{align}
\small
\label{post_mu}\vect{\mu_\ast^{\bf(s)}} &= \vect{k}_\ast^T (\vect{K}^{\bf(s)} + \sigma_{\bf s}^2\vect{I})^{-1}\vect{Y}^{\bf(s)} \\ 
\label{post_s}\vect{V_\ast^{\bf(s)}} &= k_{\ast\ast} - 
\vect{k}_\ast^T (\vect{K}^{\bf(s)} + \sigma_{\bf s}^2\vect{I})^{-1}
\vect{k}_\ast,
\end{align}
where {\small $\vect{k}_\ast = k(\vect{X}^{\bf(s)}, \vect{x}_\ast)$} and {\small $k_{\ast\ast} = k(\vect{x}_\ast, \vect{x}_\ast)$}. We use the mean of this predictive distribution for the point estimate of target outputs, denoted as $\vect{\mu}_\ast^{\bf(s)}=\vect{\mu}_\ast^{\bf(s)}(\vect{x}_\ast)$ for notational convenience. Note that we use the shared covariance function for simultaneous forecasting of the four future scores in $\vect{y}_*$. Consequently, the model assigns the same variance ($\vect{V}_{*}$) to the forecasting window. We refer to this setting as the source GP (sGP).

\subsection{Personalized GPs (pGP and tGP)}\label{pgp_tgp}
We use the notion of domain adaptive GPs (DA-GP)~\citep{liu2015bayesian,eleftheriadis2017} to personalize the population GP model to target subjects. This is achieved by sequentially adapting the GP posterior for the test subject using the data of his/her past visits, to forecast the future ADAS-Cog13 scores $\vect{y}$. This is achieved by using the obtained posterior distribution of the source (population) data as a GP prior for the GP of the ADAS-Cog13 scores of the target subject at time $t$, given by $p(\vect{y}^{\bf(t)}_{t}|\vect{x}^{\bf(t)}_t, \{\vect{X}^{\bf(s)},\vect{Y}^{\bf(s)}\}, \vect{\theta})$. Assuming that we have already observed the data of the target subject, up to time $t-1$, we use the data pairs $\{\vect{x}^{\bf(t)}_{1:t-1},\vect{y}^{\bf(t)}_{1:t-1}\}$ to correct the posterior distribution for the target subject's data at time $t$. More formally, by applying Eqs.~(\ref{post_mu}--\ref{post_s}) to $\{\vect{x}^{\bf(t)}_{1:t-1},\vect{y}^{\bf(t)}_{1:t-1}\}$, we obtain the conditional prior on the target subject data, $\mathcal{N}(\label{prior_mut}\vect{\mu}_{_{1:t-1}}^{\bf(t|s)},\label{prior_{st}}\vect{V}_{1:t-1}^{\bf(t|s)})$, where $\bf(t|s)$ denotes the conditioning order. For exact derivation, see~\citep{liu2015bayesian}. Given this prior and a test input $\vect{x}_\ast=\vect{x}_{t}^{\bf(t)}$, the correct form of the adapted posterior after observing the target subject data at visit $t$ represents the predictive distribution of the personalized GP (pGP):\\
\scalebox{1}{\parbox{1\linewidth}{%
\begin{align}
\small
\label{post_muad}\vect{\mu_{\ast}^{\bf(p)}} &= \vect{\mu_\ast^{\bf(s)}} +  {\vect{V}_{\ast}^{\bf(t|s)}}^T (\vect{V}_{1:t-1}^{\bf(t|s)} + \sigma_{\bf s}^2\vect{I})^{-1}(\vect{Y}^{\bf(t)}_{1:t-1} - \vect{\mu}_{1:t-1}^{\bf(t|s)}) \\ 
\label{post_sad}\vect{V_{\ast}^{(p)}} &= \vect{V_{\ast}^{(s)}} - 
{\vect{V}_{\ast}^{\bf (t|s)}}^T (\vect{V}_{1:t-1}^{\bf(t|s)} + \sigma_{\bf s}^2\vect{I})^{-1}
\vect{V}_{\ast}^{\bf(t|s)},
\end{align}
}}\\
where {\small $\vect{V}_{\ast}^{\bf(t|s)} = k(\vect{x}^{\bf(t)}_{1:t-1}, \vect{x}_\ast) - 
{k(\vect{X}^{\bf(s)}, \vect{x}^{\bf(t)}_{1:t-1})}^T (\vect{K}^{\bf(s)} + \sigma_{\bf s}^2\vect{I})^{-1} k(\vect{X}^{\bf(s)}, \vect{x}_\ast)$}. Eqs.~(\ref{post_muad}--\ref{post_sad}) show that final forecast by the pGP is the combination of the population-model forecast, plus a correction term.\footnote{For $t$=$1$, the population model is used instead, and from $t$=${\vect T}-4$:${\vect T}$ the future target scores are repeated to fill in the window of target scores.} The latter shifts the mean toward the distribution of the target subject and improves the model's confidence by reducing its predictive variance as more data of the target subject is observed. The inference in pGP is efficient as it uses the kernel parameters of the sGP, thus, no further training is employed.  

We also form the target-subject-specific GP (tGP) model using only the observed data of the target subject up to time $t$, i.e., the same data we used for the adaption in pGP. However, since this data set is of insufficient size to train the GP parameters -- the GP would easily overfit -- we use the kernel parameters of sGP. Unlike sGP, which has a fixed covariance matrix (trained using the source subjects), tGP continually updates its covariance matrix as more past data (up to $t$) of a target subject become available. Nevertheless, the inference procedure is the same, and for $\vect{x}_\ast=\vect{x}_t^{\bf(t)}$, the predictive distribution of tGP is given by:  

\scalebox{1}{\parbox{1\linewidth}{%
\begin{align}
\small
\label{post_mut}\vect{\mu}_{\ast}^{\bf (t)} &= {\vect{k}_{*}}^T (\vect{K}_{1:t-1}^{\bf (t)} + \sigma_s^2\vect{I})^{-1}\vect{y}^{\bf (t)}_{1:t-1} \\ 
\label{post_st}\vect{V}_{\ast}^{\bf (t)} &= k_{**} - {\vect{k}_{*}}^T (\vect{K}_{1:t-1}^{\bf (t)} + \sigma_s^2\vect{I})^{-1}
\vect{k}_{*}.
\end{align}
}}
where {\small $\vect{k}_\ast = k(\vect{x}^{\bf(t)}_{1:t-1}, \vect{x}_\ast)$}, {\small $k_{\ast\ast} = k(\vect{x}_\ast, \vect{x}_\ast)$}, and $\vect{K}_{1:t-1}^{\bf(t)}=k(\vect{x}^{\bf(t)}_{1:t-1}, \vect{x}^{\bf(t)}_{1:t-1})$. Note that the kernel matrix here increases in size after each visit of a target subject.

\subsection{Personalized GP Experts (pGPE)}
\label{pgpe}
After obtaining the pGP and tGP expert models, an optimal weighting scheme is learned to combine these two models within the proposed pGPE approach. We introduce a novel weighting scheme based on meta-learning that assigns the optimal weights to the expert models for each subject and his/her visit. More formally, recall that the predictive distribution of each expert's forecasts for the time window ahead is given by:

\begin{equation}
\text{pGP:} \,\vect{y}^{\bf(p)} \sim {\mathcal{N}}(\vect{\mu^{(p)}},\vect{V^{(p)}}), \,\,\,\,\, \text{and} \,\,\,\text{tGP:}\,\vect{y^{(t)}} \sim {\mathcal{N} }(\vect{\mu^{(t)}},\vect{V^{(t)}}).
\end{equation}
Using the expert distributions, we obtain the forecasts $\vect{y^{(g)}}$ by the pGPE model as:
\begin{equation}
\vect{y^{(g)}} = \alpha \vect{y^{(p)}} + (1 - \alpha )\vect{y^{(t)}} \sim \mathcal{N}(\alpha \vect{\mu^{(p)}} + (1 - \alpha )\vect{\mu^{(t)}},{\vect{\alpha}^2}\vect{V^{(p)}} + {(1 - \alpha )^2}\vect{V^{(t)}})
\end{equation}
In this work, we consider only the point predictions, defined as the mean of the predictive distribution defined above. To find an optimal weight $\alpha$ for each subject and his/her visit, we solve the following optimization problem:

\begin{equation}
\label{sq}
{\vect{\alpha} ^{opt}} = \mathop {\arg \min }\limits_{{\alpha _i}} \frac{1}{4N}\sum\limits_{i = 1}^N \sum\limits_{k = 1}^4 {{{\left\| {{\vect{y}^{\bf(g)}_{i,k}} - {\vect{y}_{i,k}}} \right\|}^2}},
\end{equation}
where $\vect{\alpha}^{opt}=[\alpha^{opt}_1,\dots,\alpha^{opt}_N]$, and $N=N_s\times {\vect T}$ is the total number of training samples from the training subjects and their visits. Since we seek to optimize one $\alpha$ for each data sample, Eq.\ref{sq} is a convex optimization problem with a closed form solution given by:
\begin{equation}
\label{eq:weights}
\alpha _i^{opt} = \sum\limits_{k = 1}^4\frac{\vect{y}^{\bf(g)}_{i,k} - \vect{y}_{i,k}} {\vect{y}^{\bf(p)}_{i,k} - \vect{y}^{\bf(t)}_{i,k}}\,\,\,\, , i=1\dots N.
\end{equation}
These optimal weights are used for further learning in the proposed approach. While we are able to learn these weights for the training data, it is unclear how to generate those weights for the target subjects and their visits since the future scores are unknown. To solve this, we first derive the following feature vector for each visit of the training subjects:
\begin{equation}
\vect{m}_t = [\vect{y}^{\bf(p)}_t\,\vect{y}^{\bf(t)}_t\,y_t], \,\,\dim (\vect{m}_t) = 1 \times 9,
\end{equation}
where $\vect{m}_t$ contains the four predictions per expert, pGP ($\vect{y}^{\bf(p)}$) and tGP ($\vect{y}^{\bf(t)}$), obtained by evaluating the pre-trained expert models on the data of the training subjects' visits $t=1,\dots,{\vect T}$. We also include their ground truth score for the current visit, $\vect{y}_t$. However, we do not use the data of the target subjects. Finally, the expert weights for the new data point at which we aim to forecast the future ADAS-Cog13 are obtained as:
\begin{equation}
{\alpha _*} = f({\vect{m}_*}) + {\epsilon _*},
\end{equation}
where the function $f(\cdot)$ is trained using a GP regression model with the RBF-ard kernel function and training data pairs $\{\vect{M},{\vect{\alpha} ^{opt}}\}$, where $\vect{M}=\{\vect{m}_{i=1,..,N}\}^T$. Note that the feature vector was centered by subtracting the mean of its elements. Also, we applied the quadratic expansion of such feature vector followed by the element-wise normalization with the norm of the expanded feature vector, resulting in a 55-dim feature vectors. This allowed the GP to capture the underlying relationships between these meta-features and optimal weights for the target experts. We also experimented with other feature types (e.g., using the input features, the prediction uncertainty, and the optimal weights from previous visits of the target subject), but the ones reported here achieved the best performance in the task. 
\section{Results} 
\label{exp}
\begin{figure}[!tbp]
  \centering
{\includegraphics[width=0.35\textwidth]{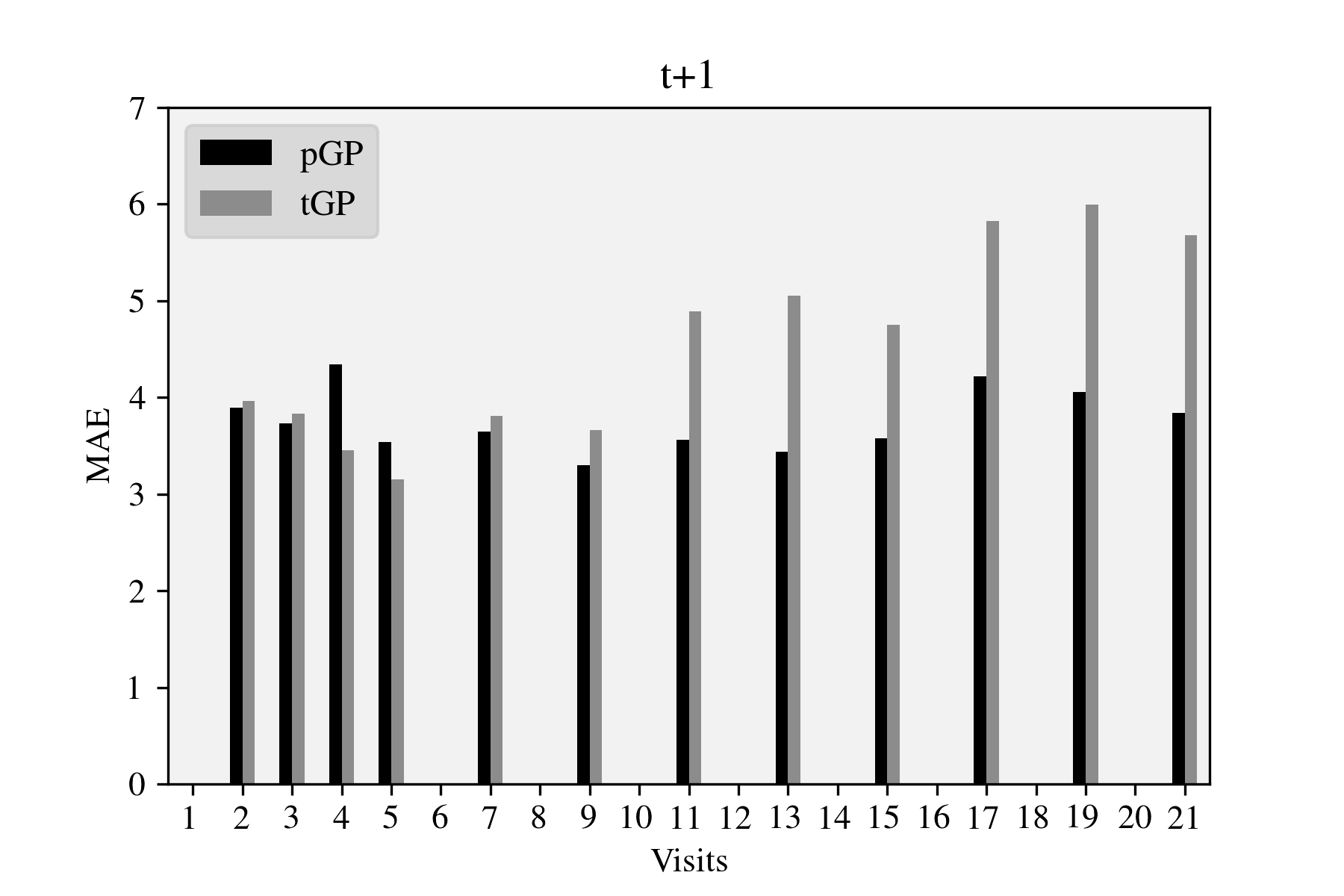}\label{fig:f1}}
  %\hfill
{\includegraphics[width=0.35\textwidth]{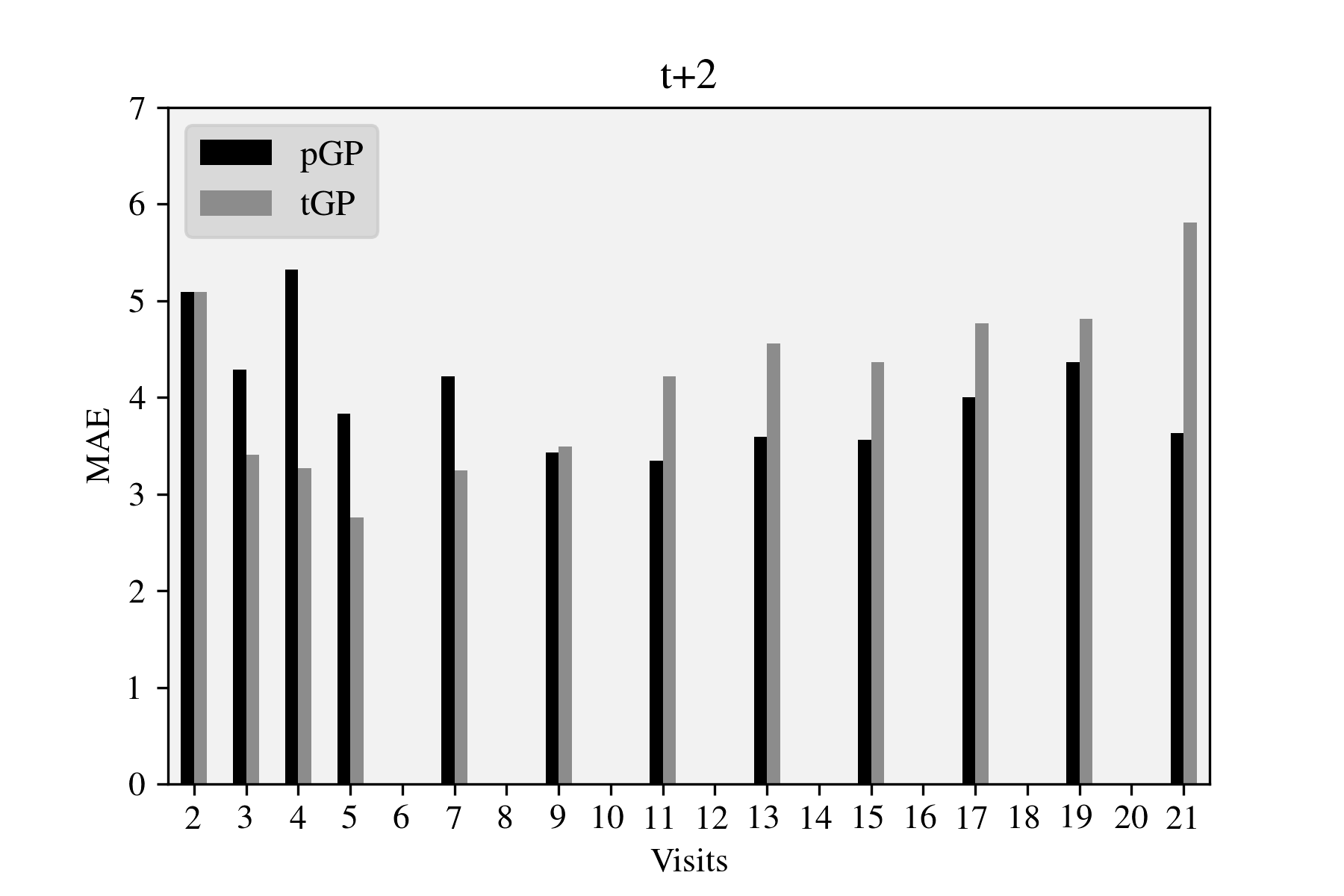}\label{fig:f2}}
   %\hfill \\
{\includegraphics[width=0.35\textwidth]{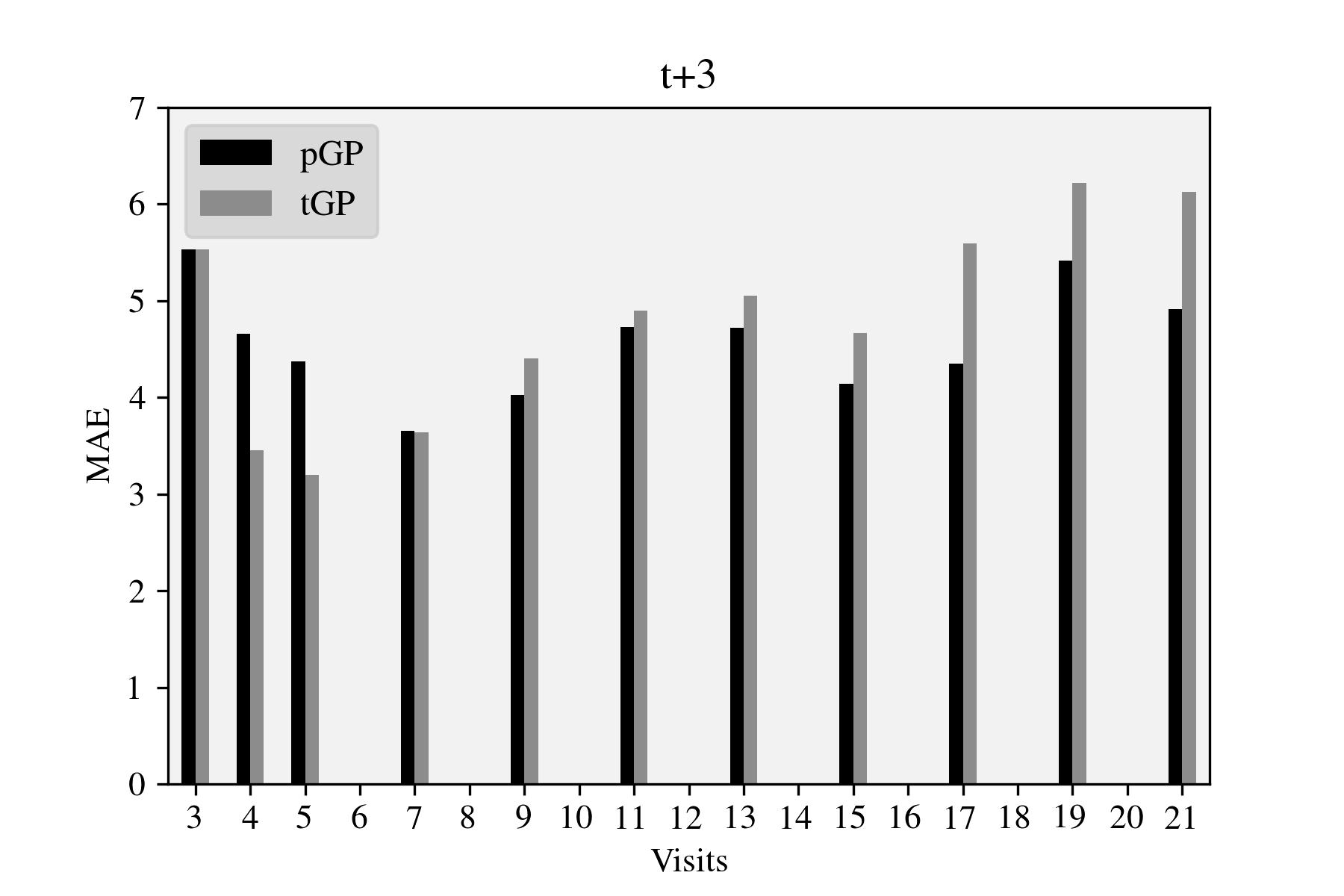}\label{fig:f3}}
  %\hfill
{\includegraphics[width=0.35\textwidth]{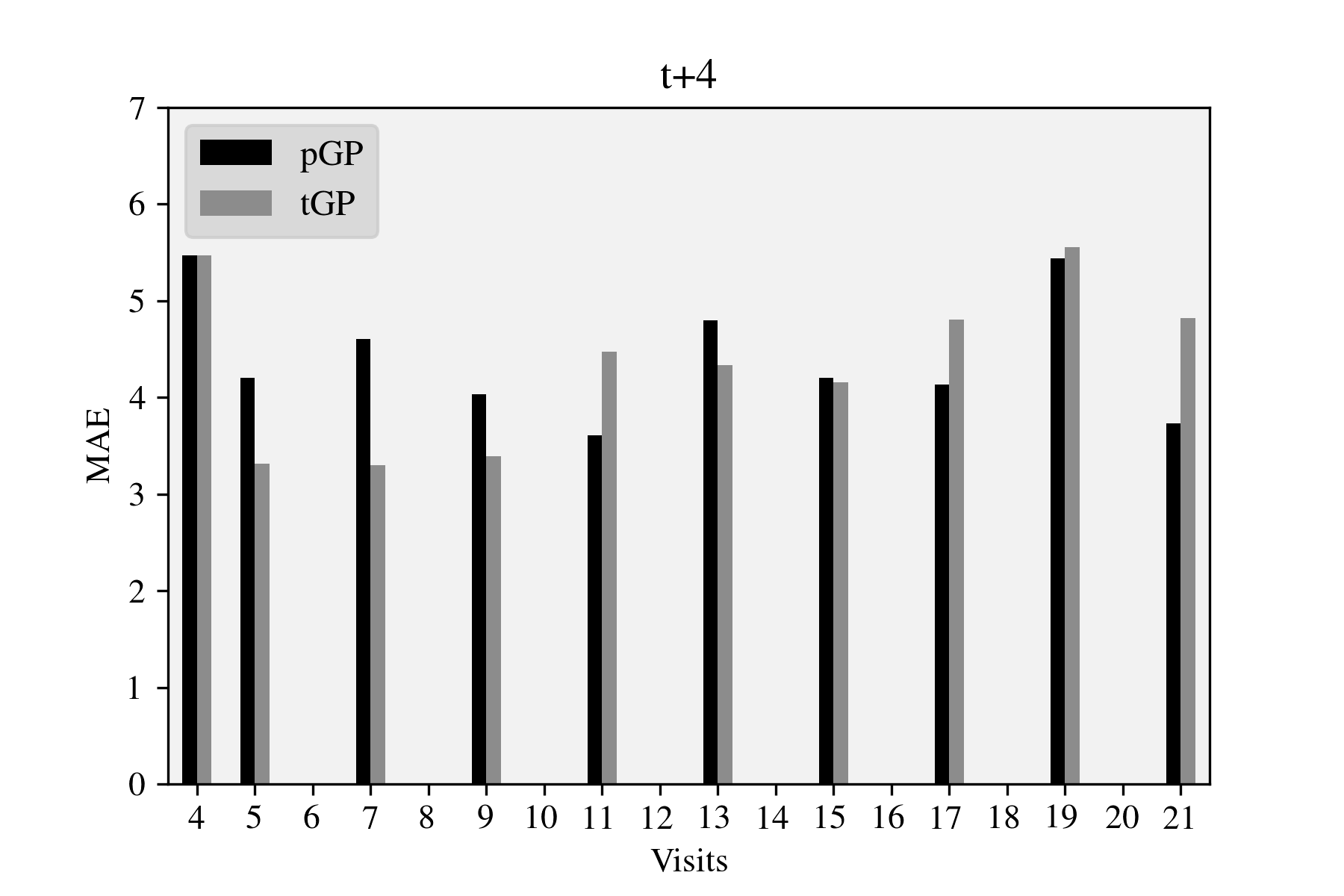}\label{fig:f4}}
  \caption{Average MAE for the pGP and tGP models computed per visit, showing specific time-steps where one of the expert models dominates over the other. This is used to derive prior weights (W$_{prior}$) for the pGPE approach by selecting the optimal model (on average) for each visit and across the four forecasting time steps. Note that after the fifth visit, the subjects in ADNI were seen on a yearly basis.}
  \label{prior}
\end{figure}

\begin{figure}[!tbp]
  \centering
{\includegraphics[width=0.2\textwidth]{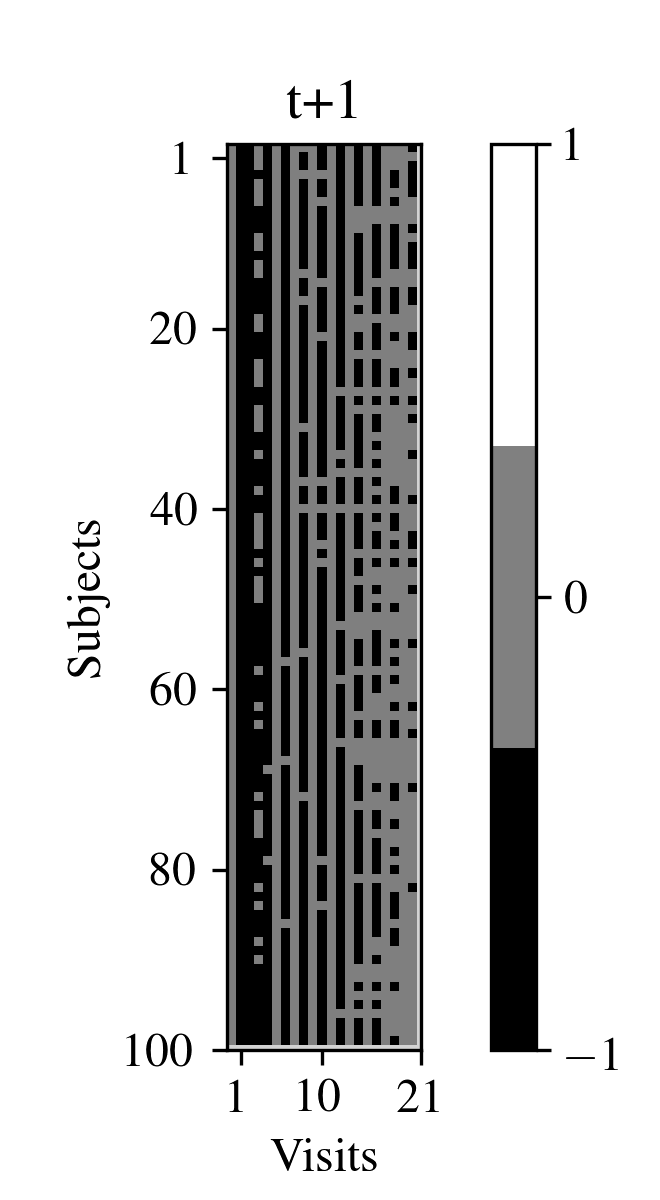}\label{fig:f1_b}}
  %\hfill
{\includegraphics[width=0.2\textwidth]{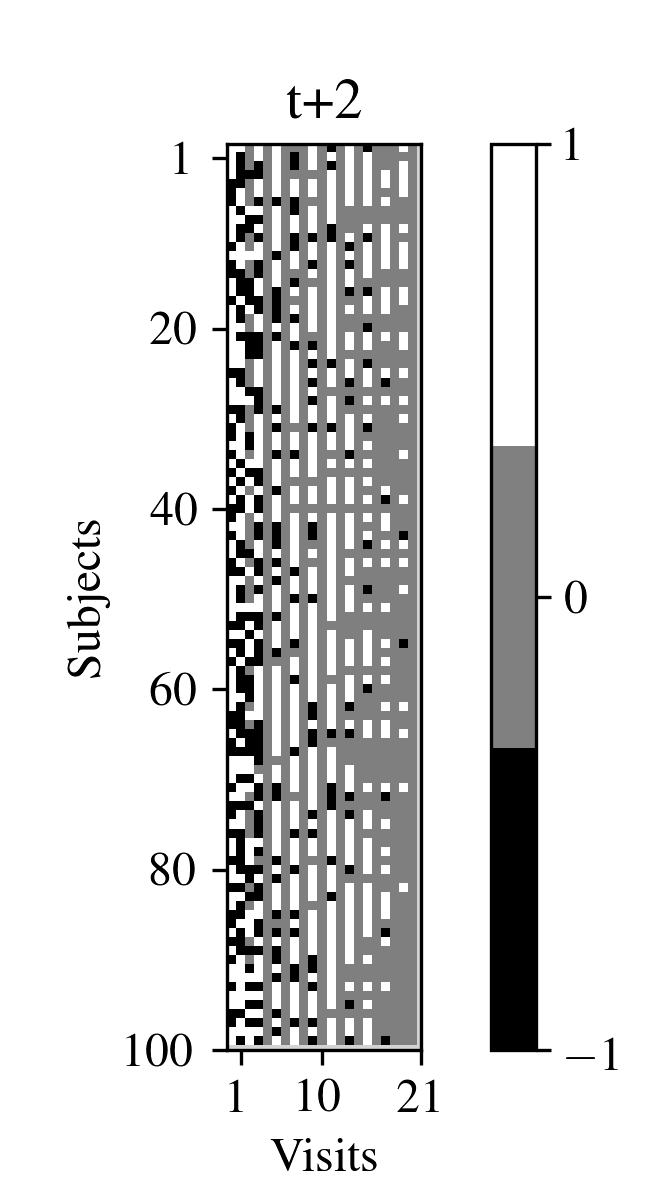}\label{fig:f2_b}}
   %\hfill
{\includegraphics[width=0.2\textwidth]{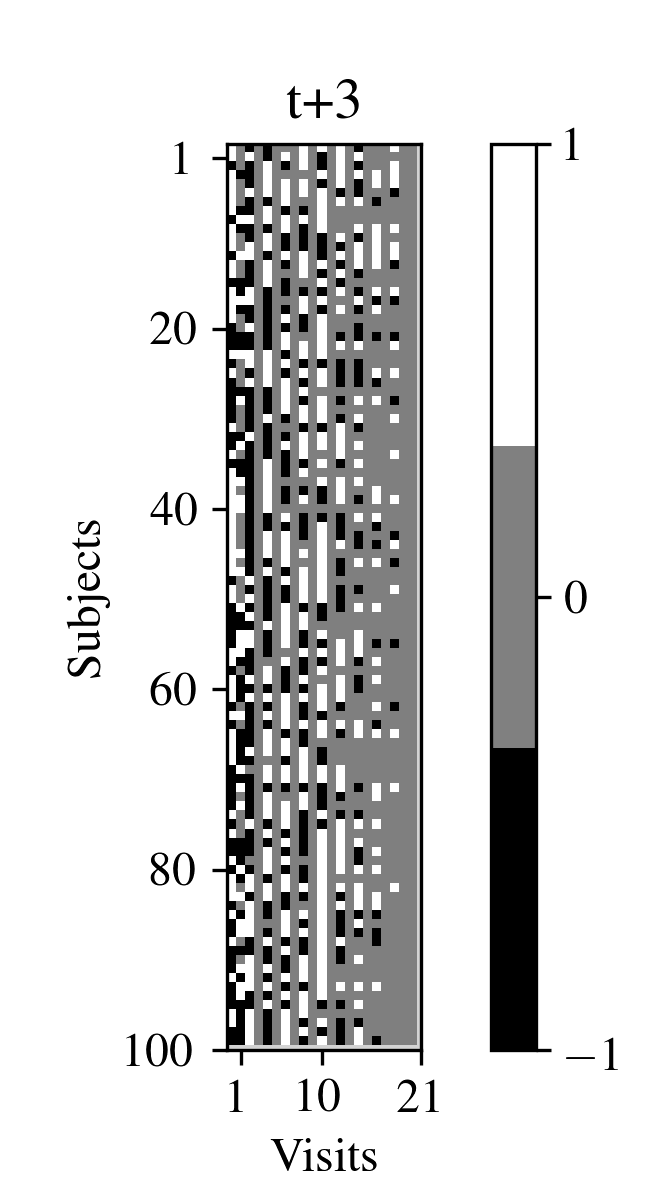}\label{fig:f3_b}}
  %\hfill
{\includegraphics[width=0.2\textwidth]{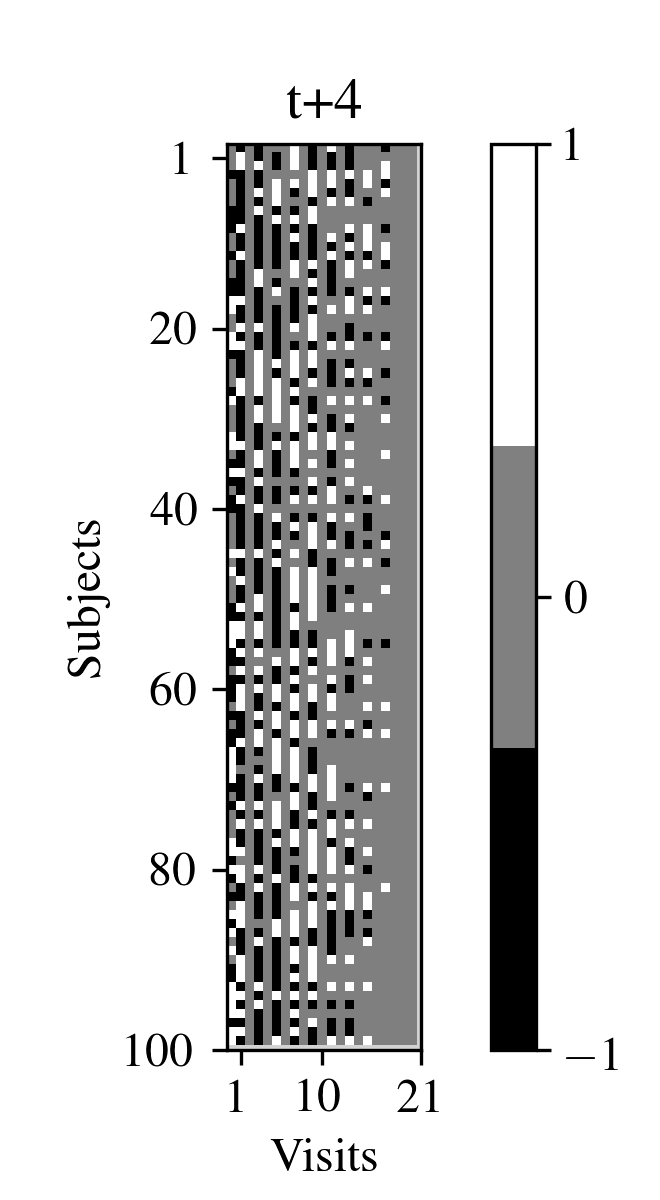}\label{fig:f4_b}}
  \caption{Plots depicting the best performing personalized model for each subject and his/her visit. For pGP, this is shown in white (-1), and for tGP this is shown in black (+1). When there is no visit data available, this is shown in gray (0). Note that the pGP always outperforms tGP at $t+1$, while for the later time steps, the models exhibit largely heterogeneous performance.}
  \label{freq}
\end{figure}

Data used in this paper come from the ADNI database \href{http://adni.loni.usc.edu/}{(adni.loni.usc.edu)}. We downloaded the standard dataset processed for the TADPOLE Challenge~\citep{tadpole2017}; this dataset represents 1,737 unique subjects and was created from the \href{https://adni.bitbucket.io/index.html}{ADNIMERGE} spreadsheet, to which regional MRI (volumes, cortical thickness, surface area), PET (FDG, AV45, AV1451), DTI (regional means of standard indices) and cerebrospinal fluid (CSF) biomarkers were added. From this data, we excluded: the PET data, because of their sparsity, the cognitive scores other than ADAS-Cog13, as well as the clinical status (CS) and normalized ventricle volumes (ICVn)\footnote{In the TADPOLE Challenge, ADAS-Cog13, CS and ICVn are treated as target outputs, so in this work we excluded the latter two from the predictors in our model.}, to construct a multi-modal feature set containing: demographics (6 features), genetics (3 features), CSF (3 features), MRI (365 features), and DTI (229 features), thus 606 input features in total. Furthermore, we selected a cohort of subjects with at least $10$ visits, and whose records were not missing more than $82.5\%$ of the subject data (from all modalities/visits), resulting in 100 subjects, 48 of whom were diagnosed with AD. We performed a 10-fold subject-independent cross-validation (i.e., each fold contained data of 10 subjects).\footnote{The models were trained using data from 9 folds, and tested on data of the remaining fold. This was repeated for each of 10 folds.} The input features were z-normalized (zero mean, unit variance). To measure the models' performance, we report $\mu\pm\sigma$ of mean absolute error (MAE) for the 10-folds. To form the forecasting window of four consecutive visits (6, 12, 18 and 24 months), we imputed the ADAS-Cog13 scores for the missing visits using the scores of the past visits. However, the reported evaluation metrics were computed using the scores for the existing visits only. We evaluated three types of GP models: (i) population-level (sGP), (ii) personalized (pGP and tGP), and (iii) personalized experts (pGPE). The settings for these models are as follows:
\begin{itemize}
\item {\bf Population-level GPs.} To forecast the future ADAS-Cog13 scores, we trained the sGP models using the input features for the visits up to time $t$ and the corresponding ADAS-Cog13 score ($y_t$). We employed two kernel functions: sGP(RBF-iso) and sGP(RBF-ard)~\citep{rasmussen2006gaussian}, where the latter learns a separate length-scale parameter for each data modality used, and $y_t$, i.e., ${\lambda _{i = 1:6}}$. We also report the performance of the sGP(RBF-iso) model when only $y_t$ from the current visit (VIS) is used as a predictor of the future scores, effectively doing the label smoothing. We denote this model as sGP(VIS).

\begin{table}
\caption{Comparison of different forecasting approaches. All the differences between pGPE and the compared models are statistically significant ($p <0.001$) based on a paired t-test with equal variances. The last row shows a lower bound on MAE, obtained when the ADAS-Cog13 scores from the {\it target} subjects are used to determine the optimal weights (W$_{opt}$) for the expert models (Sec.~\ref{pgpe}). In pGPE, the meta-weights (W$_{reg}$) for target subjects are estimated with GP tuned on {\it training} subjects.}
\label{tab_res}
\centering
\begin{center}
%\footnotesize
\begin{adjustbox}{width=0.7\textwidth}
\begin{tabular}{l|ccccc}
  \hline
    \rowcolor{Gray}
\textbf{Models}        & &&\textbf{MAE}& & \\
\hline
{}              & \text{$t+1$}& \text{$t+2$} & \text{$t+3$}& \text{$t+4$} & \text{Avg.}
                \\
\hline
\toprule   
\textbf{Base ($y_t$)} &3.74$\pm$0.17 &3.84$\pm$0.17 &4.72$\pm$0.29 &5.02$\pm$0.33 &4.34\\
\toprule      
\textbf{LassoR} &3.76$\pm$0.18 &3.85$\pm$0.16 &4.61$\pm$0.26 &4.51$\pm$0.27&4.20\\
\textbf{SVR(RBF-iso)} &3.97$\pm$0.41 &4.49$\pm$0.57 &4.88$\pm$0.49 &5.21$\pm$0.66 &4.63\\
\textbf{LSTM} &4.87$\pm$0.31 &5.44$\pm$0.36 &5.52$\pm$0.42 &6.48$\pm$0.53 &5.58\\

\toprule      
\textbf{sGP(VIS)} &3.62$\pm$0.33 &3.67$\pm$0.31 &4.37$\pm$0.54 &4.41$\pm$0.54 &4.02\\
\textbf{sGP(RBF-iso)} &3.86$\pm$0.37 &4.39$\pm$0.52 &4.72$\pm$0.44 &5.04$\pm$0.62 &4.50\\
\textbf{sGP(RBF-ard)} &3.50$\pm$0.30 &3.55$\pm$0.38 &4.11$\pm$0.61 &4.27$\pm$0.77 &3.86\\
\toprule 
\textbf{pGP(RBF-ard)} &3.50$\pm$0.31 &3.49$\pm$0.35 &4.00$\pm$0.52 &4.04$\pm$0.63 &3.76\\
\textbf{tGP(RBF-ard)} &3.70$\pm$0.37 &4.07$\pm$0.54 &3.90$\pm$0.42 &4.38$\pm$0.79 &4.01\\
\toprule 
\textbf{pGPE(W$_{prior}$)} &3.56$\pm$0.34 &3.47$\pm$0.40 &3.97$\pm$0.51 &4.04$\pm$0.65 &3.76\\
\textbf{pGPE(W$_{freq}$)} &3.70$\pm$0.37 &3.33$\pm$0.34 &3.67$\pm$0.40 & 3.82$\pm$0.59 &3.63\\             
\textbf{pGPE(W$_{ave}$)} &3.47$\pm$0.31 &3.46$\pm$0.38 &3.64$\pm$0.41 &3.85$\pm$0.60 &3.61\\
\textbf{pGPE(W$_{var}$)} &3.66$\pm$0.36 &3.98$\pm$0.51 &3.85$\pm$0.41 &4.29$\pm$0.75 &3.95\\
\textbf{pGPE(W$_{reg}$)} &{\bf 2.66$\pm$0.28} &{\bf 2.57$\pm$0.24} &{\bf 2.48$\pm$0.29} &{\bf 2.87$\pm$0.32}&{\bf 2.65}\\
\toprule  
\rowcolor{Gray}
\textbf{pGPE(W$_{opt}$)} &1.80$\pm$0.28 &1.50$\pm$0.14 &1.49$\pm$0.17 &1.54$\pm$0.37 &1.58\\
\bottomrule    
\end{tabular}
 \end{adjustbox}
\end{center}
\end{table}
\item {\bf Personalized GPs.} We also report the performance by the pGP and tGP models (Sec.~\ref{methods}). The former performs the adaptation of the sGP to the new subject using the subject data up to time $t$, while the latter builds a new kernel matrix using only the data of the new subject (however, it uses the hyper-parameters of the sGP(RBF-ard) model). We used the sGP(RBF-ard) model described above to derive the pGP and tGP models, since it achieved the best performance among the population-level GPs.  
\item {\bf Personalized GP Experts.} As described in Sec.~\ref{pgpe}, we used the pGP/tGP models as experts in the proposed pGPE approach. Below we describe different weighting schemes, including the proposed meta-weighting, we applied to these experts.\\
{\bf W$_{prior}$}: derived by inspecting the {\it average} performance of the pGP/tGP models and by applying the best performing model per visit/forecasting step (see Fig.\ref{prior}).\\
{\bf W$_{freq}$}: derived based on the frequency of pGP outperforming tGP per visit and per subject for each forecasting step. The weights range from 0 to 1, and represent the normalized sum over the columns (subjects) depicted in Fig.\ref{freq}.\\
{\bf W$_{ave}$}: derived by averaging the forecasts by pGP and tGP, as in~\citep{utsumi2018pgp}.\\
{\bf W$_{var}$}: the variance-based weighting~\citep{deisenroth2015distributed} of pGP/tGP.\\
{\bf W$_{reg}$}: the proposed regression-based meta-weighting of pGP/tGP (Sec.~\ref{pgpe})\\
{\bf W$_{opt}$}: the optimal weights obtained by applying Eq.~(\ref{eq:weights}) to the target data.
\end{itemize}
We also include the baseline comparisons with non-GP models, commonly used for forecasting tasks, and usually applied to the ADNI dataset. Specifically, we compare to Lasso Regression (LassoR), Support Vector Regression (SVR) with RBF(iso), and Long Short-Term Memory (LSTM)~\citep{hochreiter1997long}. In relation to the GP models evaluated, LassoR can be seen as our sGP with linear kernel and $l_1$ regularization, and SVR as a sparse version of our sGP. However, SVR does not provide probabilistic outputs, and, thus, it lacks a principled way for model adaptation, as done in our personalized GP models via their posterior distribution. The optimal regularization parameter and kernel width of LassoR and SVR, respectively, were selected on the held-out subjects from the training set (10/90). As input to LSTM, we tried a sliding window of the size $n=1,..,5$, containing observations from the past visits. We used again the held-out dataset to select the optimal window size ($n=2$), and the number of the LSTM output states ($h=64$). This was followed by a fully-connected linear layer ($64\times4$). During the model training, we applied a dropout (0.3) to prevent the model overfitting. Lastly, we report the results obtained by a model-free approach, denoted as Base ($y_t$), where the score from the current visit ($y_t$) was used as a forecast of the four steps ahead.

Table 1 compares the evaluation metrics for the model settings described above. By comparing different population-level GP models, we note that simple smoothing of the previous ADAS-Cog13 score ($y_t$) using sGP(VIS) is quite effective. We attribute this to the fact that the target scores from the consequent visits are highly correlated. Compared to the model free approach, Base ($y_t$), which forecasts by "carrying forward" the current score, we note that sGP(VIS) takes advantage of the GP's smoothing property, effectively leveraging the labels (i.e. ADAS-Cog13 scores) from the training population. This results in average error reduction of 0.32 over the four time steps.
However, the data-informed sGP(RBF-ard) model (that performs the smoothing of the input features) achieved the best performance overall. By learning the separate length-scale parameters for each modality, this model was able to successfully combine different data types - something that cannot be achieved with RBF-iso, resulting in the adverse performance by the sGP(RBF-iso) model. Next, we compare the personalized models (pGP and tGP) that use the sGP(RBF-ard) as the base model. It is evident that using the domain adaptive GPs to personalize the population-level model to each subject (pGP) helps to reduce MAE, the difference being more pronounced in later time steps. On the other hand, the tGP model outperforms the population-level sGP/pGP only at step $t+3$. Yet, as can be observed from Fig.~\ref{prior}, tGP outperforms these models at certain visits, even though this is not reflected in its MAE.

The bottom half of Table 1 shows results obtained from different weighting schemes that we devised to compare with our regression-based pGPE approach (pGPE(W$_{reg}$)). First, note that the weighting of the pGP/tGP models based on the heuristics derived by looking at the models' scores per-visit/subject leads to similar or better performance compared to the pGP model alone. Specifically, frequency-based weighting (W$_{freq}$) is slightly more effective than W$_{prior}$ as it is not biased by the amplitude of the per-subject MAEs when deriving the weights. This is because some subjects had much higher MAE than the others, which, evidently, did not translate well to test subjects when applying this binary model selection encoded by W$_{prior}$. Interestingly, averaging the two experts (pGP/tGP) via W$_{ave}$ further reduces MAE. As this model outperforms both the pGP and tGP alone, this also signals that the two models have very different performance per subject/visit, also evidenced by Figs.~\ref{prior}\&\ref{freq}. Thus, by simply averaging their predictions, this model reduces the prediction biases of the pGP/tGP models. We also explored the predictive variance of the pGP/tGP experts to assign different weights to the models based on their confidence in the target predictions~\citep{deisenroth2015distributed}. While this approach improved the performance of tGP alone, it hurt the overall performance of pGP. By looking into the estimated variances, we found that tGP tended to overestimate the prediction variance, due to the small amount of data (only the target subject visits' data were used to construct the kernel). 

%MAE errors per visit

%colormatrix  per subject per visit

\begin{figure}[t!]
  \centering
{\includegraphics[width=0.35\textwidth]{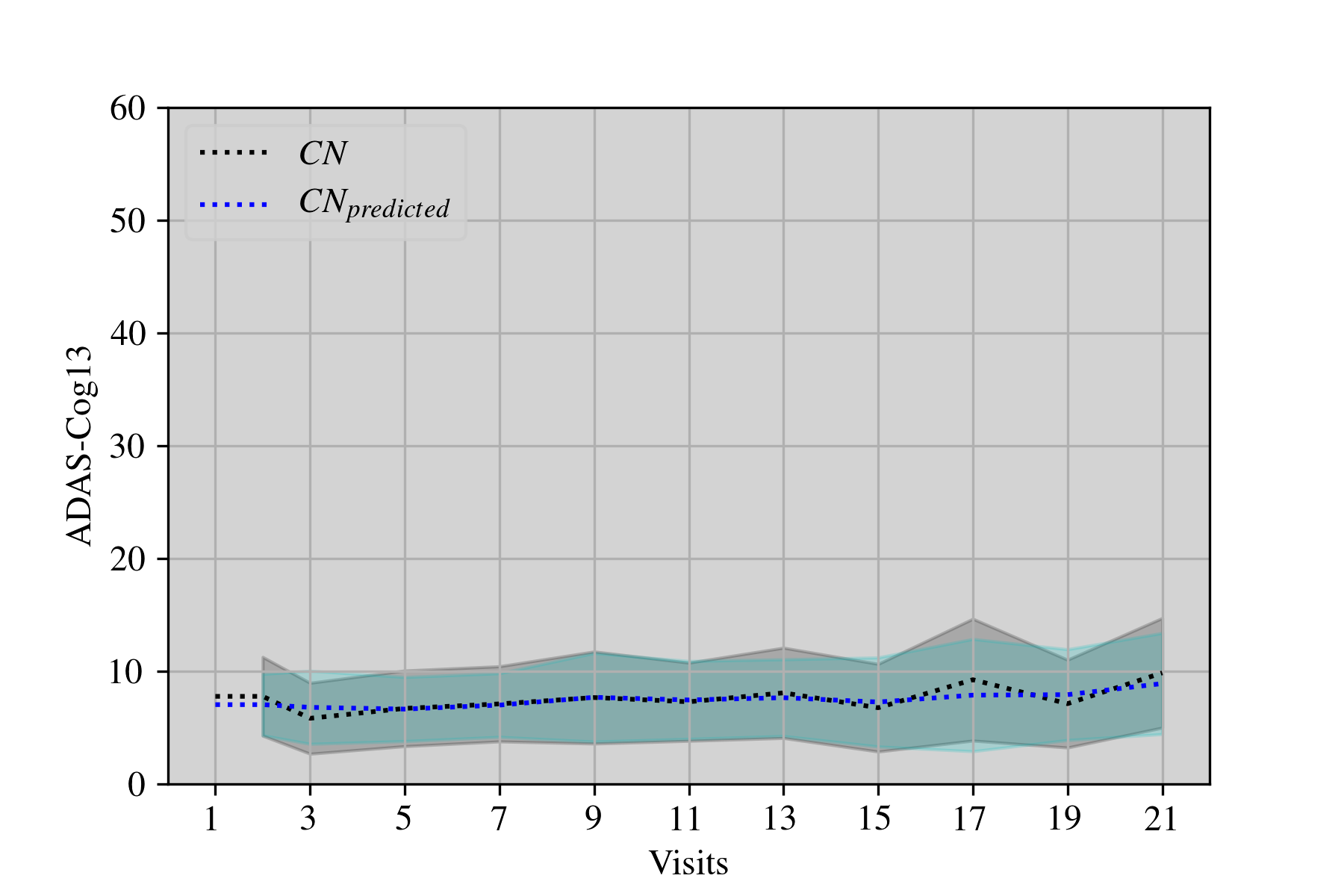}\label{fig:f1_c}}
{\includegraphics[width=0.35\textwidth]{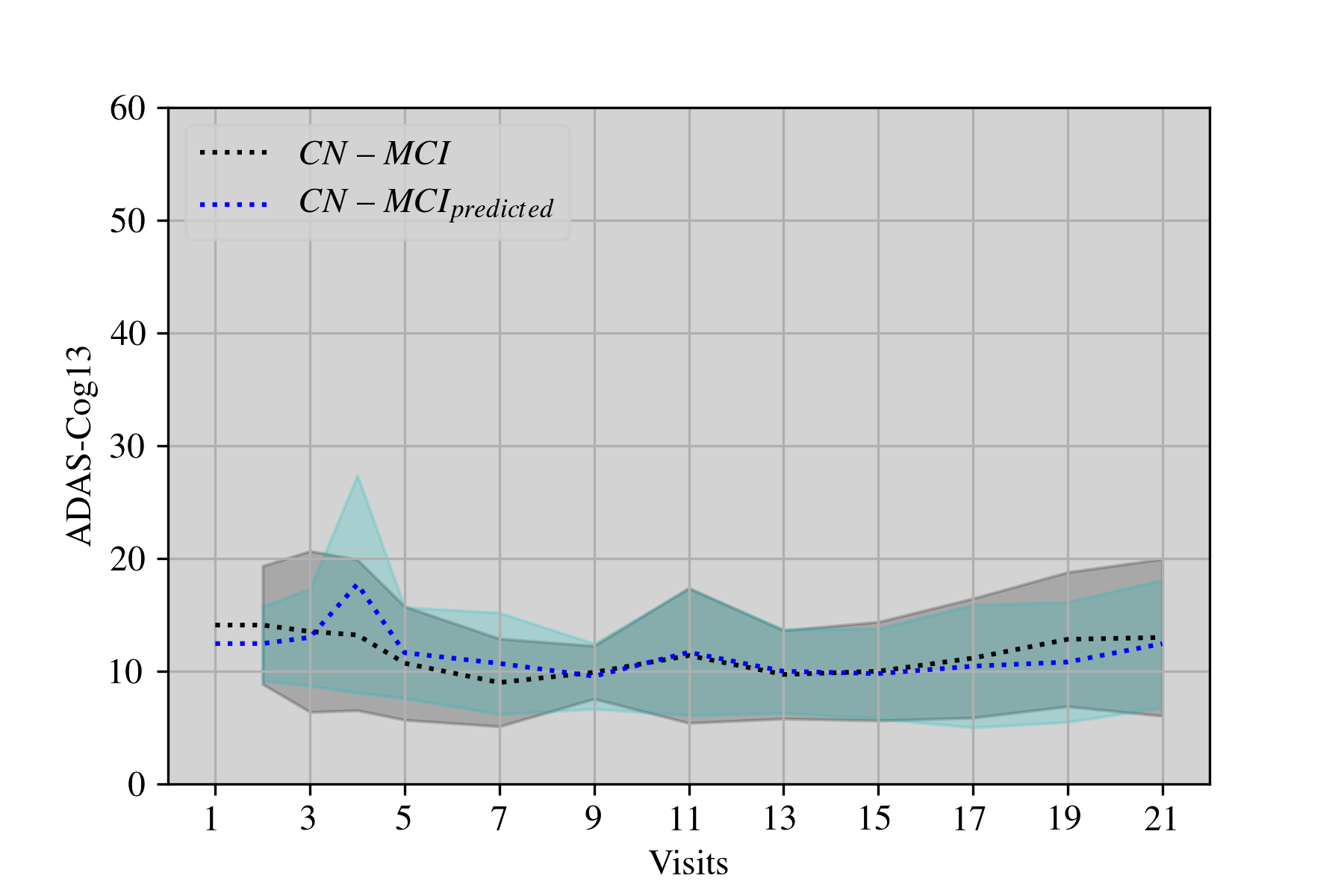}\label{fig:f2_c}} \\
{\includegraphics[width=0.35\textwidth]{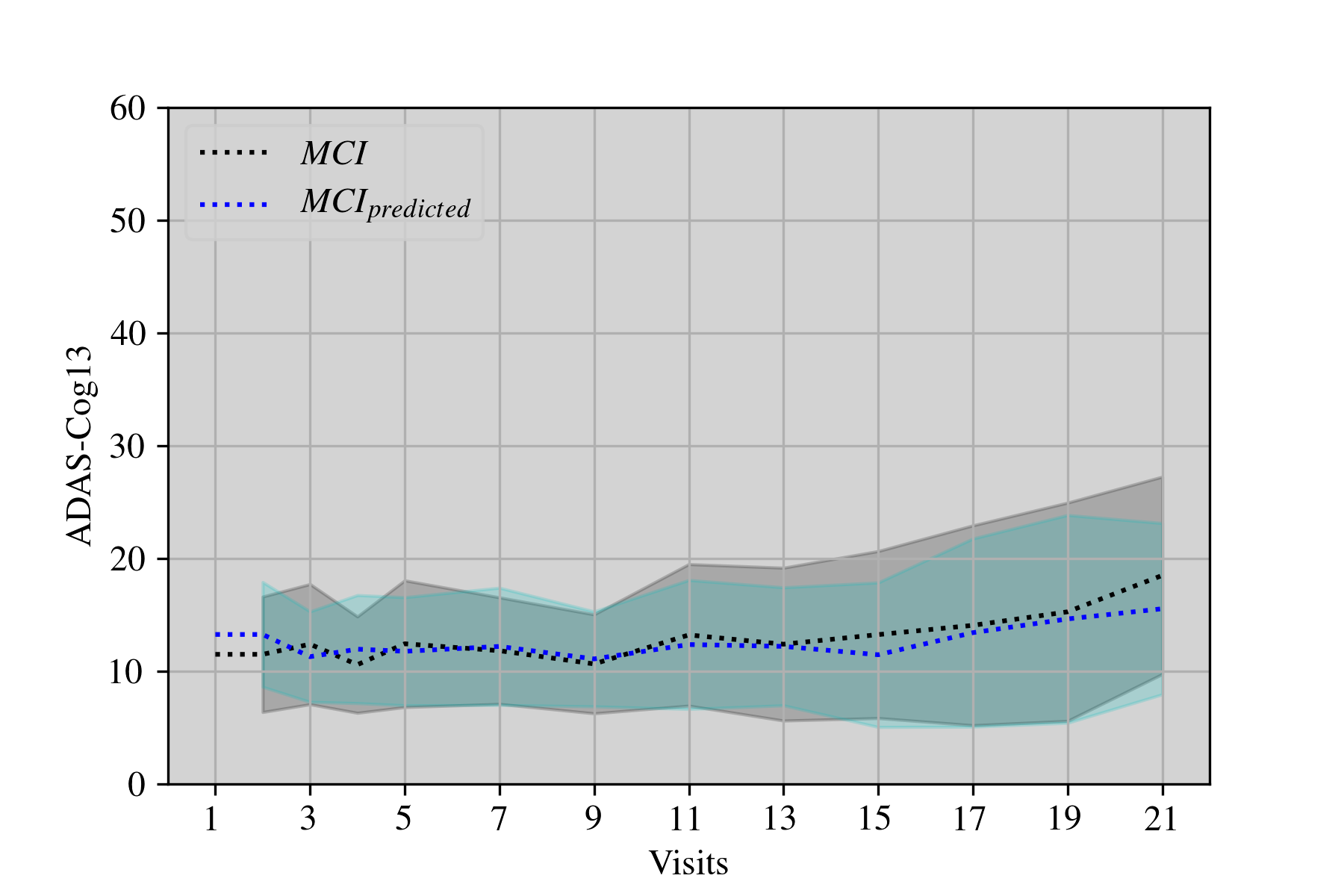}\label{fig:f3_c}}
{\includegraphics[width=0.35\textwidth]{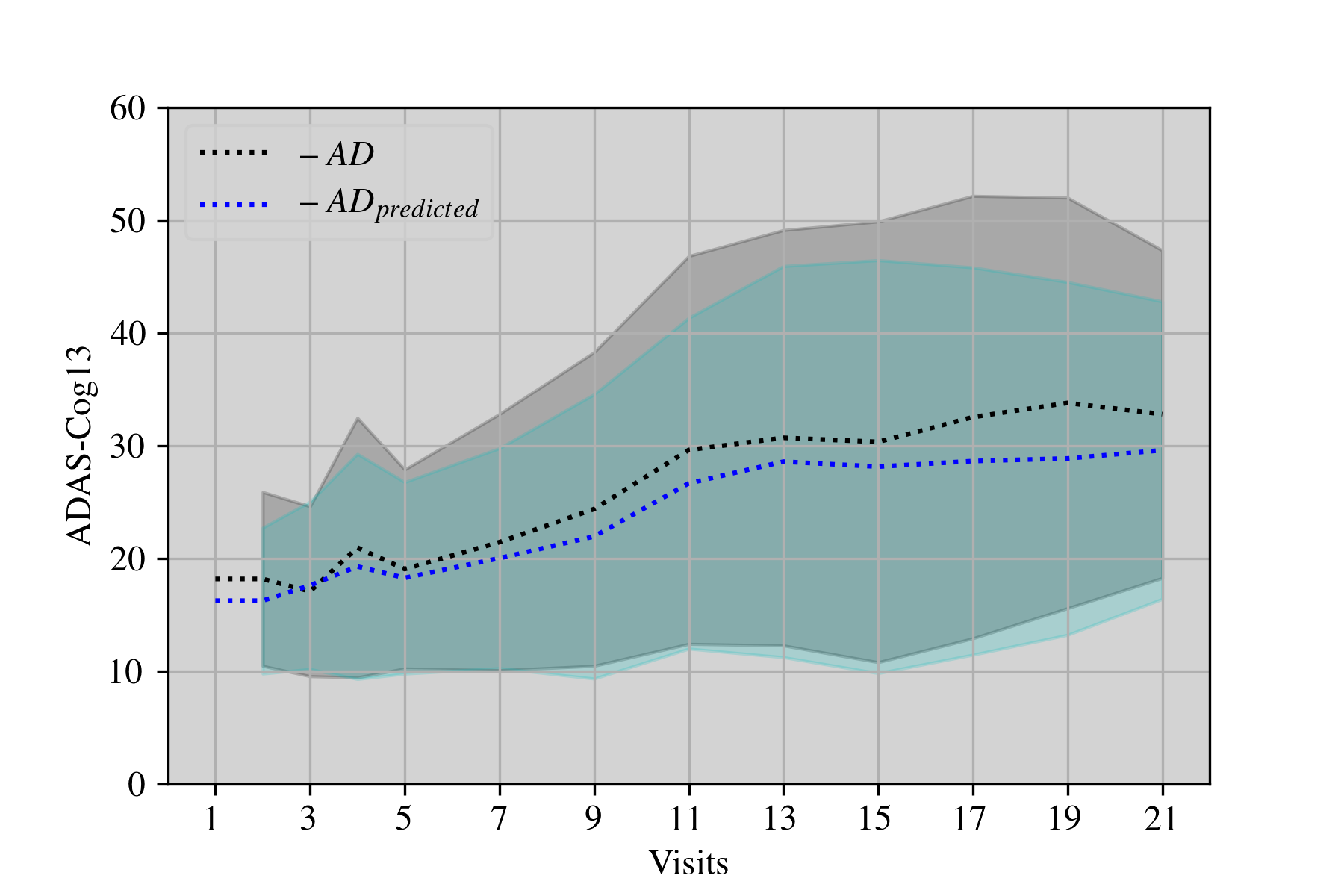}\label{fig:f4_c}}
  \caption{The graphs showing the ADAS-Cog13 scores for each visit (averaged across subjects) by the proposed regression-based pGP expert model (pGPE(W$_{reg}$)) -predicted- and ground-truth scores. Each graph shows these scores along with their standard deviation across the subjects, where the subjects were separated based on their clinical status: cognitively normal (CN), those who converted at one point from CN to mild-cognitive impairment (MCI) (CN$\rightarrow$MCI), those who stayed at MCI, and those who were diagnosed initially or converted to AD.}  
  \label{convert}
\end{figure}
The proposed regression-based weighting outperforms the compared weighting schemes by large margin, as can be seen from the per-time-step and overall results. This is consistent for the whole forecasting horizon. Note that, apart from the variance-based weighting, the other weighting approaches apply the same and fixed expert weights across all subject. This is suboptimal. The benefit of learning the subject- and visit-specific weights, as introduced here, is clear from the results obtained. This further shows that the optimal weights vary largely across subjects/visits (see Fig.~\ref{freq}); however, the proposed meta-weighting using GP was able to effectively learn those changes. To confirm this interpretation, we applied random weights sampled from the learned GP. This resulted in performance significantly lower than that obtained by the population-based models. We also included the results obtained with the optimal weights (W$_{opt}$ -- see Sec.~\ref{pgpe}) that show the lower-bound on MAE that can be achieved with the proposed pGPE approach. By looking at the performance achieved by the baselines, we note that LassoR, typically used in clinical settings because of its interpretability, outperforms the GP and SVR population models with non-linear mappings (RBF-iso). This evidences that a simple linear model can be useful in this context, however, like SVR that is commonly used in prediction tasks in ADNI, it does not provide a probabilistic output. On the other hand, due to the probabilistic nature of GPs, pGP can easily adapt to the target subjects, which result in lower average MAE. While LSTMs have shown great performance in various clinical time series forecasting tasks, in our experiments on the ADNI data, it underperformed. We attribute this to the highly heterogenous and irregularly sampled data, which, evidently, negatively affects the ability of LSTM to leverage its memory properties.    

Fig.~\ref{convert} shows the graphs depicting the predicted ADAS-Cog13 scores for each visit (averaged across subjects) by the proposed pGPE(W$_{reg}$) that we plot against the ground-truth scores. Each graph shows these scores along with their one standard deviation across the subjects, where the subjects were separated based on their clinical status (not used as input in the models): 25 cognitively normal (CN), 20 who converted at one point from CN to mild-cognitive impairment (MCI) (CN$\rightarrow$MCI), 7 who stayed at MCI, and 48 who were diagnosed initially or converted to Alzheimer's Disease (AD) in one of their visits. Note that the proposed approach can accurately forecast the trend in changes of the target scores. As can be seen, the subjects who were diagnosed with AD or converted to AD at one of the visits have a large deviation from the ADAS-Cog13 scores for the other three types of clinical status, with this difference being more pronounced towards later visits, as expected. While we show here that the proposed approach matches well the distributions of the cognitive scores at the sub-group level, the proposed approach can be used to predict the trend of the cognitive changes for each target subject, and associate it with future changes in his/her clinical status (e.g. to detect subjects that will rapidly progress over a period of 2 years). We plan to investigate the feasibility of such approach in our future work.  

%\end{figure}

% % here I will include per patient results
% \begin{figure*}[!tbp]
%   \centering
% {\includegraphics[width=0.48\textwidth]{images/pp1.png}\label{fig:f4}}
%   \hfill
% {\includegraphics[width=0.48\textwidth]{images/pp2.png}\label{fig:f5}}
%   \caption{Each subject's average MAE average over the forecasting window $w$. We contrast the joint pGP+tGP model against pGP (a), tGP (b), and sGP (c), respectively. The subjects are sorted based on their decreasing differences to better visualize the individual performances.}
% \end{figure*}

\section{Conclusions}
\label{headings}
%! Please verify edits I made here for accuracy:
Accurate forecasting of changes in the biomarkers of AD is challenging, mainly because of the highly heterogeneous, noisy and missing nature of the clinical data available for this task, but also due to the highly pronounced individual differences in such data. As a step forward, this work developed a novel personalized approach for accurate forecasting of ADAS-Cog13, a significant biomarker of AD, up to two years in the future. The proposed personalized GP framework for time series forecasting uses the notion of meta learning to determine an optimal weighting scheme for individual expert models based on personalized GP models. We showed on a sub-cohort of ADNI subjects that this approach brings significant improvements over population-level GP models. Furthermore, we showed that the proposed personalized experts alone cannot generalize well enough across all the subjects and their visits. By contrast, the introduced personalized meta-weighted GP framework enables accurate forecasting of changes in ADAS-Cog13, outperforming the existing personalized GP models, and traditional models commonly used for forecasting of time-series clinical data. We also showed that the newly introduced meta-weighting is more effective than the standard weighting schemes for GP expert models. In the future, we plan to extend our meta-weighted GPE framework so that it can handle more than two experts and forecast more than one biomarker of AD simultaneously. For this, we will investigate the notion of multi-task GPs and reinforcement learning, for actively learning to select the optimal experts for each subject/visit. We also plan to employ techniques for filling in the missing data in order to increase the cohort of the subjects from ADNI for evaluation of our models. This automated approach for forecasting of cognitive changes in AD could augment and assist clinicians by providing them with intelligent data summarization and decision support tools for early identification of at-risk subjects and construction of informative clinical trials.

\small{
%\bibliography{sample}
% use bibtex file with first names abbreviated:
\bibliography{reformatted_bib}
}
%\appendix
%\section*{Appendix A.}

\end{document}